\documentclass{article}

\usepackage{microtype}
\usepackage{graphicx}
\usepackage{booktabs} 

\usepackage{hyperref}

\usepackage[accepted]{tom2023}

\usepackage{amsmath}
\usepackage{amssymb}
\usepackage{mathtools}
\usepackage{amsthm}

\usepackage{enumitem}       
\usepackage{float}          
\usepackage{caption}        
\usepackage{subcaption}     
\usepackage{tikz}           

\usepackage{ragged2e}       

\usepackage{inconsolata}    
\usepackage[T1]{fontenc}

\usepackage[capitalize,noabbrev]{cleveref}

\theoremstyle{plain}

\theoremstyle{definition}

\theoremstyle{remark}

\usepackage[textsize=tiny]{todonotes}

\icmltitlerunning{Inferring the Goals of Communicating Agents from Actions and Instructions}

\begin{document}

\twocolumn[
\icmltitle{Inferring the Goals of Communicating Agents \\ from Actions and Instructions}



\icmlsetsymbol{equal}{*}

\begin{icmlauthorlist}
\icmlauthor{Lance Ying}{equal,harvard,mit}
\icmlauthor{Tan Zhi-Xuan}{equal,mit}
\icmlauthor{Vikash K. Mansinghka}{mit}
\icmlauthor{Joshua B. Tenenbaum}{mit}
\end{icmlauthorlist}

\icmlaffiliation{harvard}{School of Engineering and Applied Sciences, Harvard University, Cambridge, MA, USA}
\icmlaffiliation{mit}{Department of Brain and Cognitive Sciences, Massachusetts Institute of Technology, Cambridge, MA, USA}

\icmlcorrespondingauthor{Lance Ying}{lanceying@seas.harvard.edu}
\icmlcorrespondingauthor{Tan Zhi-Xuan}{xuan@mit.edu}

\icmlkeywords{Theory of Mind, ToM}

\vskip 0.3in
]



\printAffiliationsAndNotice{\icmlEqualContribution} 

\begin{abstract}
When humans cooperate, they frequently coordinate their activity through both verbal communication and non-verbal actions, using this information to infer a shared goal and plan. How can we model this inferential ability? In this paper, we introduce a model of a cooperative team where one agent, the principal, may communicate natural language instructions about their shared plan to another agent, the assistant, using GPT-3 as a likelihood function for instruction utterances. We then show how a third person observer can infer the team's goal via multi-modal Bayesian inverse planning from actions and instructions, computing the posterior distribution over goals under the assumption that agents will act and communicate rationally to achieve them. We evaluate this approach by comparing it with human goal inferences in a multi-agent gridworld, finding that our model's inferences closely correlate with human judgments $(R = 0.96)$. When compared to inference from actions alone, we also find that instructions lead to more rapid and less uncertain goal inference, highlighting the importance of verbal communication for cooperative agents.
\vspace{-12pt}
\end{abstract}

\section{Introduction}

Human cooperation is a flexible, interactive process that involves the mutual observation and interchange of a great variety of signals and cues, providing information about the goals, intentions, beliefs, and other mental states of the people involved. Some of these signals are implicit, such as goal-directed actions, whereas others are explicit, such as verbal communication. In order to navigate cooperative life, social agents like ourselves must integrate this multiplicity of information into coherent theories of others' minds, drawing inferences about shared or individual goals and plans that can serve as guides to cooperative action.

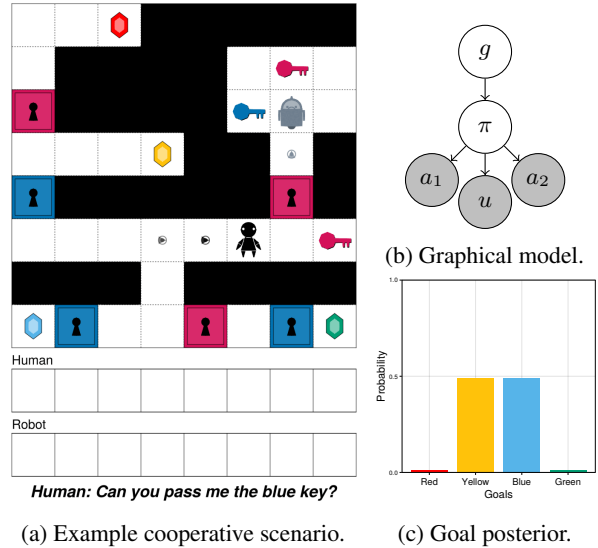
\begin{figure}[t]

\centering

\begin{subfigure}[b]{0.575\columnwidth}
\includegraphics[width=\textwidth]{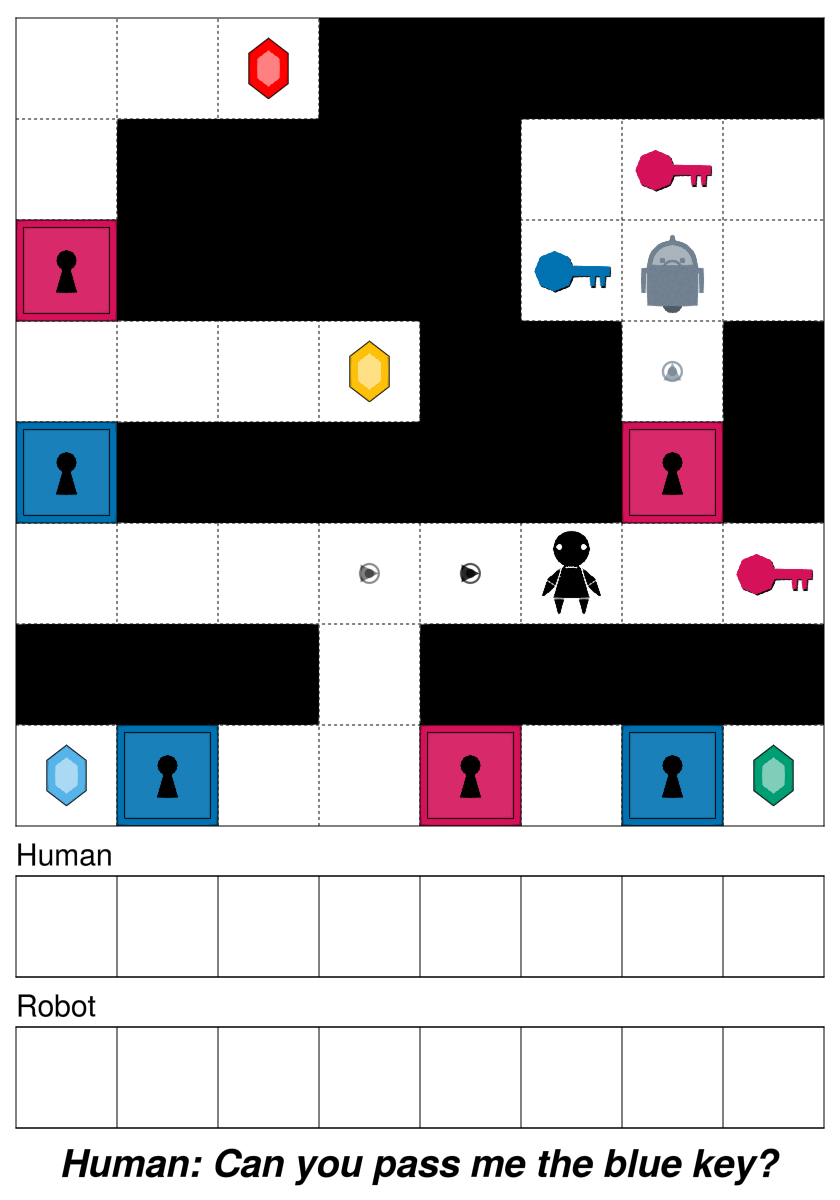}
\caption{Example cooperative scenario.}
\end{subfigure}
\begin{subfigure}[b]{0.375\columnwidth}
\centering
\begin{subfigure}[b]{\textwidth}
\centering
\begin{tikzpicture}[every node/.style={draw, circle, minimum size=2em, font=\footnotesize}]
  \node[draw, circle] (g) {$g$};
  \node[draw, circle, below of=g] (pi) {$\pi$};
  \node[draw, circle, below left of=pi, fill=lightgray] (a1) {$a_1$};
  \node[draw, circle, below of=pi, fill=lightgray] (u) {$u$};
  \node[draw, circle, below right of=pi, fill=lightgray] (a2) {$a_2$};

  \path[->] (g) edge (pi);
  \path[->] (pi) edge (a1);
  \path[->] (pi) edge (u);
  \path[->] (pi) edge (a2);
\end{tikzpicture}
\caption{Graphical model.}
\end{subfigure}

\begin{subfigure}[b]{\textwidth}
\includegraphics[width=\textwidth]{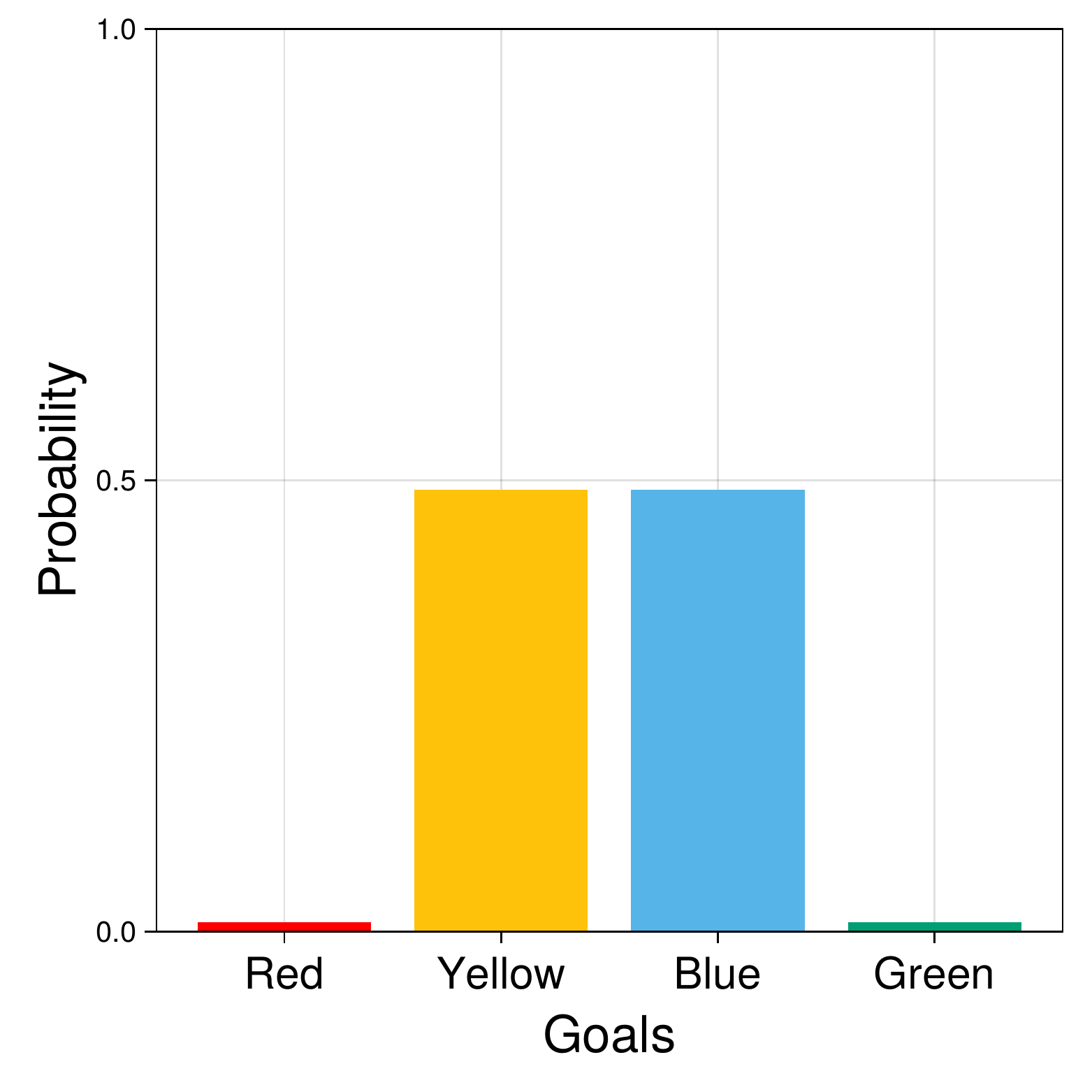}
\caption{Goal posterior.}
\end{subfigure}
\end{subfigure}
\caption{An overview of our framework: (a) A human-robot team cooperates to achieve a shared goal $g$ (one of the 4 colored gems), with the human uttering an instruction $u =$ \textit{"Can you pass me the blue key?"} to the robot assistant. (b) We model the team as forming a \emph{joint} plan $\pi$ to achieve their goal $g$, which dictates their actions $a_1$ (human), $a_2$ (robot). The human also communicates part of this plan as an instruction $u$. (c) Bayesian inverse planning produces a posterior distribution over goals $p(g|u,a_1,a_2)$.}
\label{fig:overview}
\vspace{-12pt}
\end{figure}

What is it that explains this inferential ability in humans, and how can it inform the design of cooperative AI systems? We take steps toward an answer by building upon a long tradition in cognitive science that models human linguistic and action understanding as processes of Bayesian interpretation: On one hand, Bayesian theory-of-mind (BToM) posits that humans understand other's actions by inferring the goals and beliefs that explain those actions as rational \cite{baker2009action, baker2017rational}. On the other hand, rational speech act (RSA) theory suggests that humans interpret other's utterances not just in terms of bare semantics, but also the pragmatic intentions they imply \cite{goodman2013knowledge,goodman2016pragmatic}. Since each of these frameworks are formulated in terms of Bayesian inference over the mental states that might explain observed actions or instructions respectively, it is natural to combine them, achieving \emph{joint} inference from actions \emph{and} uttered instructions.

In this paper, we develop a Bayesian model of communicating team agents that incorporates aspects of both these frameworks. The team consists of two agents, a principal (played by a human) who may communicate instructions to an assistant (played by a robot), both of whom act in order to achieve a shared goal (illustrated in Figure \ref{fig:overview}(a)). Unlike related work that explores how the assistant should infer the principal's goal \cite{hadfield2016cooperative,jeon2020reward,squire2015grounding}, our task is to infer the \emph{team's} goal given their actions and communicated instructions, producing a distribution over goals (Figure \ref{fig:overview}(c)).

To do so, we follow recent work in cooperative agency by modeling the team as a \textbf{group agent}, bypassing the challenge of recursive mental reasoning \cite{shum2019theory, tang2020bootstrapping, tang2022exploring, wu2021too}. We implement this model as a \textbf{probabilistic program} that comprises a goal prior, joint planner, and utterance model (schematically depicted in Figure \ref{fig:overview}(b)), extending a line of research that uses the flexibility of probabilistic programming to modularly specify agent models in terms of deterministic, probabilistic, and black-box components \cite{owain2017agent,cusumano2017probabilistic,seaman2018nested,zhi2020online,berke2020learning}. This in turn allows us to easily integrate \textbf{neural language models as flexible utterance likelihoods} given hypothesized goals and plans, building upon the insight made by \citet{lew2020leveraging} and subsequent papers \cite{dohan2022language, li2023lampp} that (large) language models (LLMs) can be used as modular components in larger probabilistic models.

To evaluate this model, we conduct a series of computational and human experiments that tested the model's ability to accurately infer the goal of a team in a multi-agent gridworld environment, and also how well it explains the goal inferences of third-person human observers when they are provided with the same actions and instructions. For comparison, we also perform experiments in a baseline setting where instructions are omitted, allowing us to isolate the role that language information plays in goal inference. We find that human goal inferences are highly and robustly correlated with the inferences produced by our model, and that language instructions greatly accelerate the convergence of inferences to the true goal, with remaining ambiguity resolved by action information. Collectively, these findings suggest that our model is a viable explanation for how humans infer goals from actions and instructions, as well as a promising route towards building communicative AI assistants that act on the basis of well-calibrated goal inferences.

\section{Modeling Communicating Agents}

In accordance with the principle of rational action \cite{gergely2003teleological,baker2009action} and rational speech act theory \cite{goodman2016pragmatic}, we model communicative cooperators as rational agents who act and communicate efficiently to achieve shared goals. However, a complete model of rational Bayesian communication and action requires a great deal of sophistication: Since each agent in the cooperating team may not initially know the team's goal (as is the case for our robot assistant), a third person observer would have to model not only the team's shared goal, but also each agent's \emph{beliefs} about their shared goal, including how those beliefs are formed through goal inference. In addition, agents who know the goal (such as our human principal) would have to be modeled as \emph{pedagogically} selecting utterances in order to best reduce listeners' uncertainty about the shared goal and plan \cite{shafto2014rational}. As a further level of sophistication, an observer might model an assistive agent as a \emph{pragmatic listener} who reasons about what a pedagogical speaker might utter \cite{fisac2017pragmatic}.

\subsection{Communicating Teams as Group Agents}

We sidestep these multiple levels of recursive reasoning by opting to model a cooperating team as \emph{a single group agent}: Instead of separately representing the mental states of both the human principal and the robot assistant, we model them as a singular mind with a shared goal $g$, and a joint plan $\pi$. Given this joint plan $\pi$, the principal agent utters an instruction $u$ to communicate the plan, and each agent $i$ takes actions $a_{i,t}$ at timestep $t$ according to the plan (a simplified graphical model is shown in Figure \ref{fig:overview}b):
\begin{alignat}{3}
&\textit{Goal Prior:} \qquad &g &\sim P(g) \label{eq:goal-prior} \\
&\textit{Joint Planning:} \qquad &\pi &\sim P(\pi|g) \label{eq:joint-planning} \\
&\textit{Utterance Model:} \qquad &u &\sim P(u|\pi) \label{eq:utterance-model} \\
&\textit{Action Selection:} \qquad &a_{1,t}, a_{2,t} &\sim P(a_{1,t},a_{2,t} | \pi) \label{eq:action-selection}
\end{alignat}
A peculiar aspect of this model is that the principal is assumed to communicate the plan $\pi$ through an utterance $u$, despite both agents supposedly having shared mental states: Why communicate the plan, if everybody knows what it is?

Nonetheless, there are good reasons for using this as a model for the purposes of goal inference. First, since our group agent model is much simpler than the complete model described earlier, it can serve as a \emph{resource-rational approximation} \cite{lieder2020resource} of the true dynamics, allowing third-person observers to infer the goals of cooperating teams while avoiding the need to represent the mental states of individual agents (as in \citet{shum2019theory}), or assume pedagogical communication (as in \citet{milli2017should}). Second, our model is highly plausible as an \emph{Imagined We} model from the perspective of an assistive agent who is part of team. In the Imagined We (IW) framework, cooperative agents avoid excess recursive reasoning by imagining themselves as a group agent with a shared goal, albeit a goal that may be unknown to individual members \cite{tang2020bootstrapping,tang2022exploring}. To act, group members infer their shared goal by asking, "What is it that \emph{we} want, that best explains our actions so far?" They then direct their actions towards the inferred goal, resulting in decentralized goal convergence. Building upon recent work that applies the IW framework to pragmatic communication \cite{stacy2021modeling}, our model effectively extends the listener component of the IW framework to account for joint communication and action: To infer a shared goal, assistive agents ask, "What is that \emph{we} want, that best explains our actions \emph{and} instructions?"

\subsection{Model Components}

\begin{figure}[t]
    \centering
    \begin{subfigure}[b]{\columnwidth}
    \algrenewcommand\algorithmicprocedure{\textbf{model}}
    \begin{algorithmic}
        \Procedure{utterance-model}{$\pi$}
        \State \textbf{parameters}: $p_\text{communicate}, \mathcal{E}$
        \State $a^*_{1:t} \gets \textsc{rollout-policy}(\pi)$
        \State $\alpha_{1:k} \gets \textsc{extract-salient-actions}(a^*_{1:t})$
        \State $p \gets p_\text{communicate} \textbf{ if } (k > 0) \textbf{ else } (1 - p_\text{communicate})$
        \State $c \sim \textsc{bernoulli}(p)$
        \If{$c = \textsc{true}$}
            \State $u \sim \textsc{language-model}(\alpha_{1:k}, \mathcal{E})$
        \EndIf
        \EndProcedure
    \end{algorithmic}
    \caption{Utterance model $P(u, c| \pi)$ as a probabilistic program}
    \end{subfigure}

    \vspace{5pt}
    \begin{subfigure}[b]{\columnwidth}
    \scriptsize
    \begin{tabular}{rp{0.79\columnwidth}}
    \textbf{Input:} & \RaggedRight{\texttt{(handover robot human key2) where (iscolor key2 blue)}} \\
    \textbf{Output:} & Hand me the blue key. \\
    \textbf{Input:} & \RaggedRight{\texttt{(unlockr robot key1 door1) where (iscolor door1 red)}} \\
    \textbf{Output:} & Can you unlock the red door for me? \\
    \textbf{Input:} & \RaggedRight{\texttt{(handover robot human key1) (handover robot human key2) where (iscolor key1 green) (iscolor key2 red)}} \\
    \textbf{Output:} & Can you pass me the green and the red key? \\
    \end{tabular}
    \caption{Paired examples $\mathcal{E}$ of salient actions $\alpha_{1:k}$ and utterances $u$}
    \end{subfigure}
    \caption{Our utterance model is a probabilistic program (a) that extracts salient actions $\alpha_{1:k}$ from a joint plan $\pi$, then samples an utterance $u$ using a language model (in our case, GPT-3 \textsc{Curie}) given $\alpha_{1:k}$ and few-shot examples $\mathcal{E}$ in its prompt. Several examples are shown in (b).}
    \label{fig:utterance-model}
    \vspace{-12pt}
\end{figure}

Having defined the high level structure of our model, we now describe its individual components. For the goal prior, $P(g)$, we use a uniform distribution over a fixed set of possible goals $g \in G$. In the context of our environment (Figure \ref{fig:overview}a), a goal $g$ corresponds to the human picking up one of the four colored gems.

To model joint planning in an efficient manner, we make the assumption that agent's actions are \emph{ordered} --- i.e., the agents take turns, with the principal (human) acting at each step $t$ while the assistant waits, before the assistant (robot) acts at $t+1$ while the principal waits. This limits the branching factor of planning, while preserving the optimal solution \cite{boutilier1996planning}. Under this assumption, we model joint planning as the process of computing a joint Boltzmann policy $\pi$ for a goal $g$:
\begin{equation}
    \pi(a_{i,t} | s_t, g) = \frac{\exp{\frac{1}{T} Q_g^*(s_t, a_{i,t})}}{\sum_{a_{i,t}'} \exp{\frac{1}{T} Q_g^*(s_t, a_{i,t}')}}
\end{equation}
where $s_t$ is the current state, $a_{i, t}$ is the action taken by agent $i$ at $s_t$, $T$ is a temperature parameter, and $Q_g^*(s_t, a_{i,t})$ is the (negated) cost of the optimal plan from $s_t$ to goal $g$ with $a_{i, t}$ as its first action. This models a team that is \emph{noisily optimal} in how it acts, with the amount of noise controlled by $T$. Importantly, $Q_g^*(s_t, a_{i,t})$ need not be computed in advance, but can instead be computed online for each action $a_{i,t}$ and state $s_t$ observed during the inference. We do this using real-time adaptive A* search as an incremental shortest-path planner \cite{koenig2006real}, avoiding the prohibitive cost of computing a $Q$-value for every state and action via value iteration (used by related inverse reinforcement learning algorithms, e.g. \citet{ramachandran2007bayesian, ziebart2008maximum}), while using the $Q$-values computed by previous A* searches to inform future searches.

With the policy $\pi$ computed, we model action selection by sampling actions according to the policy. In addition, we can use $\pi$ to model the instruction $u$ that the principal agent utters according to the following process:

\begin{enumerate}[topsep=2pt, itemsep=2pt]
    \item Rollout the policy $\pi$ with temperature $T = 0$ to get an optimal sequence of actions $a^*_{1:t}$ to the goal.

    \item Extract \emph{salient} actions $\alpha_{1:k}$ from $a^*_{1:t}$ to be communicated to the assistive agent.

    \item Generate a natural language instruction or request that communicates the salient actions $\alpha_{1:k}$ (or avoid communicating if there are none).
\end{enumerate}

We implement the above process as a probabilistic program that combines deterministic, stochastic and neural components, shown in Figure \ref{fig:utterance-model}(a). Steps 1 and 2 are deterministic, with step 2 implemented by filtering out non-salient actions like directional movement, and keeping only important actions for the assistant to perform, such as handing over keys or unlocking doors. This approximates a pragmatic speaker in the RSA framework \cite{goodman2008rational}, communicating instructions that trade-off informativeness and utterance cost by mentioning only the most relevant actions to achieving the team's shared goal. Step 3 has two parts: (i) if there are $k > 0$ salient actions to communicate, the program decides with high probability to communicate an utterance, with this choice denoted by $c$; (ii) if this occurs (i.e. $c = \textsc{true}$), then the utterance $u$ is generated using a neural language model, conditioned on both the salient actions $\alpha_{1:k}$ and a series of few-shot examples $\mathcal{E}$ that are included in the prompt (Figure \ref{fig:utterance-model}(b)). In our implementation, we use the \textsc{Curie} variant of GPT-3 \cite{brown2020language} to serve as the utterance likelihood $P(u| \alpha_{1:k}, \mathcal{E})$, since we found it reasonably calibrated when evaluating the probability of an utterance $u$, and did not require the more realistic forward-generation abilities of larger language models. However, any language model that defines a probability distribution over string tokens can in principle be used.
  
\subsection{Goal Inference from Actions and Instructions}

Using the model described above, our aim is to compute the posterior distribution over goals $g$ given an instruction $u$, whether an instruction was communicated $c$\footnote{Note that $c$ must be true if $u$ is observed.}, and a series of actions $a_{i,1:t}$ for each agent $i$:
\begin{multline}
    P(g|u, c, a_{1, 1:t}, a_{2, 1:t}) \propto P(g, u, c, a_{1, 1:t}, a_{2, 1:t}) = \\
    P(g) P(u, c| \pi_g) \textstyle{\prod}_{\tau=1}^{t} P(a_{1, \tau}| \pi_g) P(a_{2, \tau} | \pi_g)   
\end{multline}

Since all the terms in the joint distribution  can be computed exactly\footnote{Since $\pi$ is deterministic given $g$ in our model, we omit $P(\pi | g)$ from the expression and replace $\pi$ with $\pi_g$.}, we can perform exact Bayesian inference by updating the unnormalized weights $w_t^g$ for each goal $g$ as new evidence arrives, then normalizing the weights to get the probability $P_t^g$ for goal $g$ at timestep $t$:
\begin{align*}
    w_0^g &\gets P(g) P(u, c| g) \\
    w_t^g &\gets w_{t-1}^g P(a_{1, \tau}| \pi_g) P(a_{2, \tau} | \pi_g) \\
    P_t^g &\gets w_t^g / \textstyle{\sum}_{g'}(w_t^{g'})
\end{align*}

We implement this inference algorithm as an exact variant of Sequential Inverse Plan Search \cite{zhi2020online} using the particle filtering extension of the Gen probabilistic programming system \cite{cusumano2019gen, Zhi-Xuan_GenParticleFilters_jl}, which can be configured to support exact inference by disabling random sampling, while still automating all the necessary weight updates.

\section{Experiments}

\begin{figure*}[ht]
    \centering
    \includegraphics[width=\textwidth]{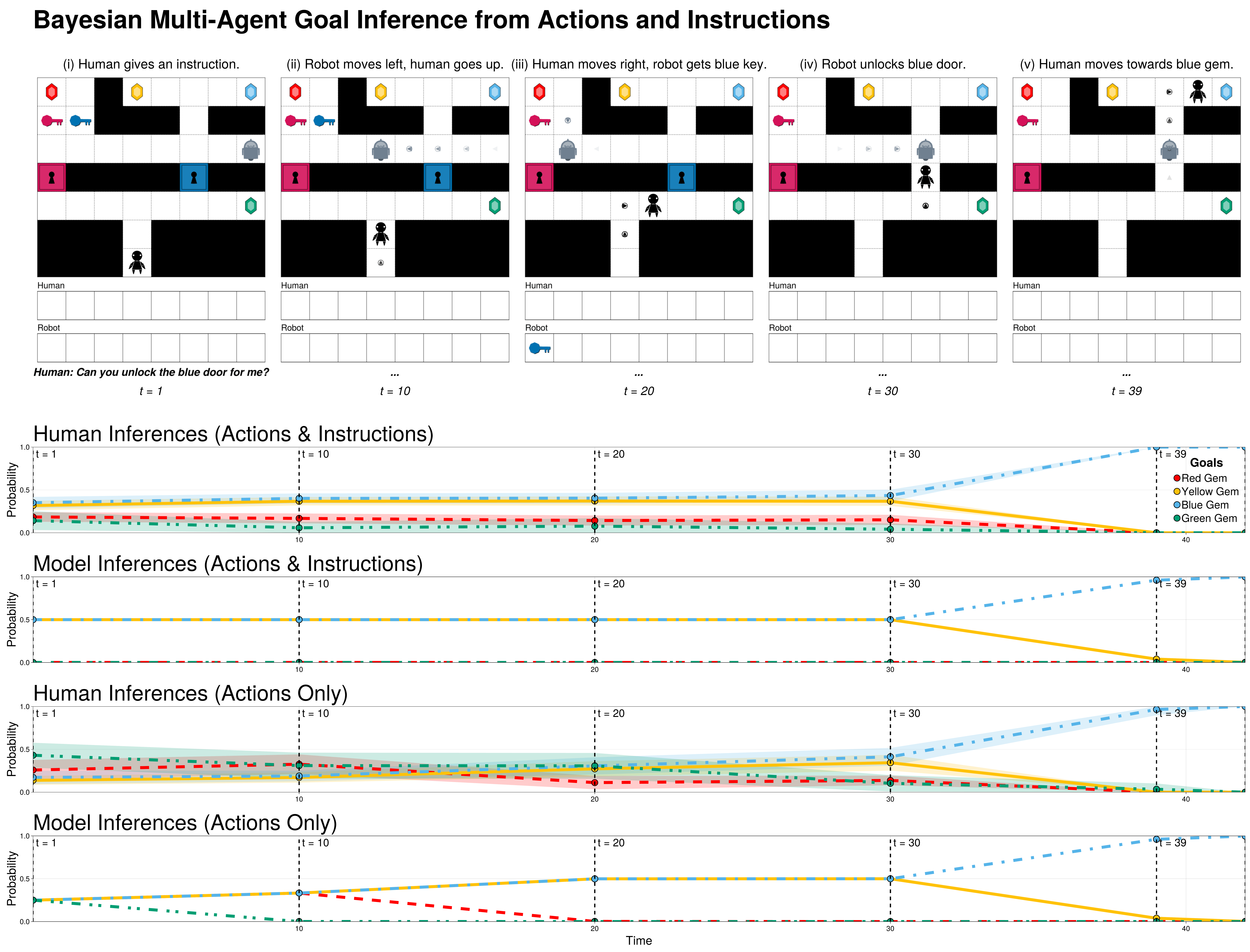}
    \caption{Goal inferences over time from the actions and instructions of a human-robot team, where the team's goal is for the human to pick up one of the four colored gems. \emph{(Row 1)} Frames from an illustrative state-action trajectory $(s_{1:t}, a_{1:t})$, with the initial utterance $u =$ \textit{"Can you unlock the blue door for me?"} shown below the first frame. \emph{(Row 2)} Average human inferences (w. 95\% CI) given both the instruction and actions, elicited at the selected frames. \emph{(Row 3)} Model inferences via Bayesian inverse planning from instructions and actions. \emph{(Row 4)} Average human inferences (w. 95\% CI) given actions only, without any instructions provided. \emph{(Row 5)} Model inferences via Bayesian inverse planning from actions only.}
    \label{fig:storyboard}
\end{figure*}

To evaluate both the scientific validity and performance of our model, we conducted a human and computational experiments, comparing our model's goal inferences against goal inferences elicited from humans. As a baseline, we used an "Actions Only" model that does not model or condition upon uttered instructions. Below we describe the environment we used to conduct our experiments, the dataset of action and instruction stimuli we generated, followed by the human experiment and model fitting procedures.

\subsection{Environment Description}

In order to study goal inference in a multi-agent setting, we adapt the Doors, Keys, \& Gems gridworld from \citet{zhi2020online} into a multi-agent environment, where a human principal and a robot assistant collaborate to retrieve a target gem (Figure \ref{fig:overview}(a)). In this environment, agents may need to pick up keys and unlock doors so as to reach desired items. Keys and doors are colored, and agents can only unlock a door using a key of the same color, after which the key is exhausted. To allow for cooperative behavior, agents may pass held items to each other if they are on adjacent grid cells. In addition, the robot is not allowed to pick up gems, reflecting its role as an assistive agent. If they have nothing useful to do, agents may also wait at their current location, which carries 60\% the cost of other actions.

\subsection{Dataset Generation}

We constructed 6 instances of the multi-agent Doors, Keys, \& Gems environment with varying maze designs and item locations. In each of these environment instances, we created 2--4 action sequences from the initial state to a goal gem, generated through a combination of automated planning and manual modification in order to increase the diversity and goal ambiguity associated with each action sequence. For each action sequence, we wrote a natural language instruction that the human might communicate to the robot in a variety of styles (e.g. requests like \textit{"Can you unlock the red and blue door for me?"}, or commands like \textit{"Pass me the blue key."}). Instructions are all communicated at the beginning of each action sequence. In total, we constructed 20 stimuli of action trajectories paired with instruction utterances, reflecting a range of cooperative scenarios.

\subsection{Human Experiment Design}

Our human experiment involved two experimental conditions in order the isolate the effect of language information on goal inference: (i) with-instructions and (ii) without-instructions condition. In the with-instructions condition, participants were shown animated trajectories of the human-robot team as stimuli, along with the instructions given by the human to the robot at the initial state (see Figure \ref{fig:storyboard} for selected frames). In the without-instructions condition, the participants were shown the same animated stimuli, but without the instruction. Animated trajectories were segmented at certain judgment points, and participants provided their goal inferences at each judgment point by selecting all gems they thought were likely to be the team's goal (see Appendix). We inserted 4--5 judgment points for each trajectory, depending on the length of the trajectory.

\subsection{Participants}

We recruited 120 US participants fluent in English via Prolific (age 19--62 with mean 34; 49 women, 67 men, 4 non-binary), 60 of whom were assigned to each of the experimental conditions. Participants were paid at a rate of 15 USD per hour. Each participant provided goal inferences for 10 out of the 20 stimuli. In total, each stimulus was rated by approximately 30 participants. 

Before viewing the stimuli, participants went through a tutorial and answered five comprehension questions, which all participants passed. Participants also earned points proportional to their Brier skill score, a measure of well-calibrated prediction \cite{weigel2007discrete}, and were paid a bonus for earned points (\$1 to \$3 per participant on average) to incentivize good-faith effort at inferring the goal. One participant in the with-instruction condition was excluded from our analysis due to an low total point score below the first quartile (Q1) minus the inter-quartile range (IQR).

\begin{figure*}[t]
    \centering
    \includegraphics[width=0.98\textwidth]{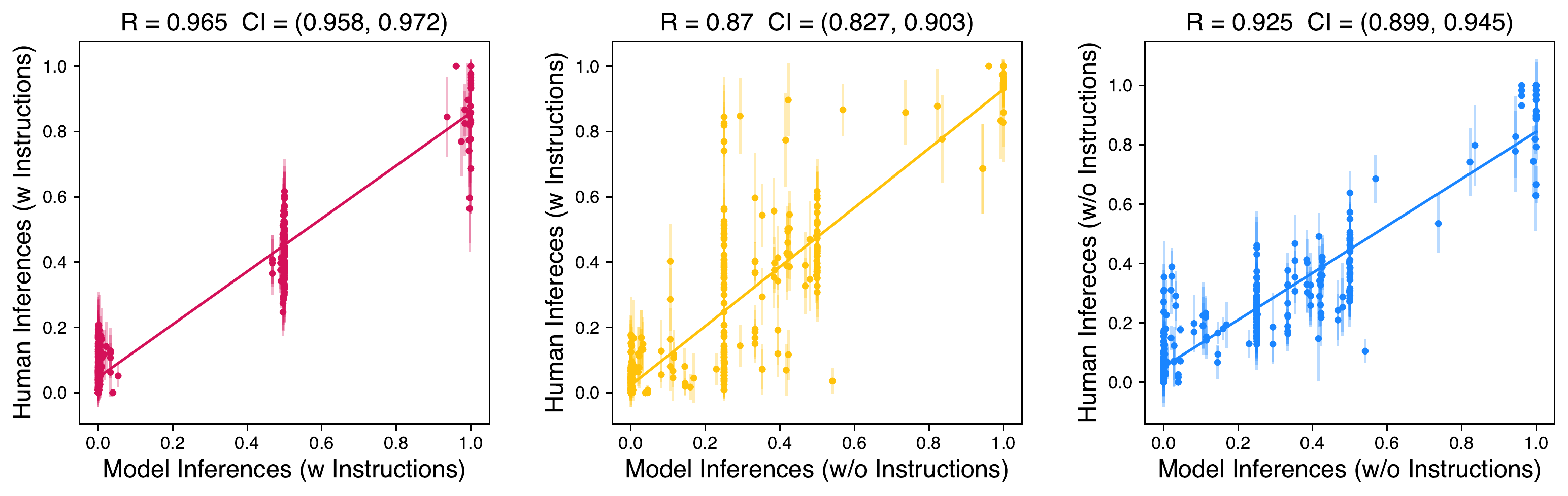}
    \caption{Correlation plots between average human inferences and model inferences, in the with-instructions condition \textit{(left)} and without-instructions condition \textit{(right)}. Our model displays a good fit in both conditions. In contrast, the without-instructions model correlates less well with human inferences from both actions and instructions \textit{(middle)}.}
    \label{fig:correlation}
\end{figure*}

\subsection{Computational Experiments and Model Fitting}

We ran Bayesian goal inference with our model on the same set of stimuli we provided to humans. We fixed the probability of communicating an utterance for plans with salient actions to $p_\text{communicate} = 0.95$, and also fixed the set of 7 few-shot examples $\mathcal{E}$ used by our utterance model, leaving no other free parameters besides the Boltzmann policy's temperature $T$, which we varied from 0.0625 to 16 in powers of two. Our model was run on both the stimuli including the instructions and without the instructions, with the latter serving as a baseline model that only computes goal inferences from action observations. To fit the model, we computed Pearson's correlation coefficient $R$ between model inferences at each judgment point vs. average human inferences at the same judgment point. We found that $T = 1.0$ led to high correlation in both conditions, achieving the highest geometric mean of $R$ across conditions. As such, the following results all use $T = 1.0$ unless stated otherwise.

\subsection{Results}

\paragraph{Qualitative Analysis}
Figure \ref{fig:storyboard} shows an illustrative example of multi-agent goal inference from actions and instructions over time. In this example, the human principal is able to directly reach the green gem, but the other three gems, along with the robot, are locked behind a red door and a blue door. As such, the human principal requires the robot's help to unlock one of the doors in order to reach one of those gems. The red gem is closer to the red door, while the yellow and blue gems are closer to the blue door. Without instructions (i.e. observing actions only), both humans and our model place a (close-to)-uniform prior over the four gems. But with instructions, both our model and humans place higher probability on the yellow and blue gems at $t = 1$, while down-weighting the red and green gems. This is because a rational agent would not utter an instruction for the green gem, and would instruct the robot to unlock the red instead of blue door if the goal was the red gem. 

As the example progresses through $t = 10$ (robot moves toward the key) and $t = 20$ (robot picks up the blue key), the team's actions gradually provide information about the goal, such that the action-only model eventually stops considering the green gem and red gem as live possibilities. Humans are slightly more uncertain when shown only actions, down-weighting the red gem, but maintaining the possibility that the green gem might be the goal even at $t = 20$. In contrast, this goal uncertainty is considerably reduced when instructions are provided at the start, with lower uncertainty in humans manifesting not just as faster convergence to the true goal, but also smaller confidence intervals (reflecting lower population variance). Only at $t = 30$ (robot unlocks the door) do human inferences from actions alone converge towards human inferences with language instructions. From this point onward, the team's actions have revealed enough information that goal inferences are the same regardless of the initial information provided by language.

\paragraph{Correlational Analysis}
To quantitatively evaluate the fit between our model's inferences and human participants' inferences, we run a correlational analysis and show the results across the two experimental conditions in Figure \ref{fig:correlation}. As can be seen, our model's goal inferences are strongly correlated with human judgments in both experimental conditions, with a Pearson's $R$ of 0.965 (95\% CI of [0.958, 0.972]) in the with-instructions condition (left plot), and a Pearson's $R$ of 0.925 (95\% CI of [0.899, 0.945]) in the without-instructions condition (right plot). The 95\% confidence intervals are calculated through bootstrapping with 1000 samples.

Figure \ref{fig:correlation} also shows that the correlation coefficient is higher in the with-instructions condition. We suspect that this is because language instructions reduce the uncertainty and variance in human inferences, leading to a better fit. The left-most plot of Figure \ref{fig:correlation} also shows that the probability ratings form small clusters around values of 0, 0.5, and 1, whereas the data points in the without-instruction condition have a much wider spread. This indicates that instructions help the observer effectively reduce the set of possible goals to just two or one goals. As a baseline comparison, we plot the correlation between our model's inferences without instructions and human inferences with instructions (middle plot). This resulted in a lower correlation coefficient of 0.87, with a 95\% CI of [0.827, 0.903], demonstrating the importance of modeling inference from instructions to explain human goal inferences.

\begin{table*}[t]
    \centering
    \footnotesize
    \begin{tabular}{lcccccc}
    \hline
        & \multicolumn{2}{c}{\textbf{t = first}} & \multicolumn{2}{c}{\textbf{t = median}} & \multicolumn{2}{c}{\textbf{t = last}} \\
          \cline{2-7}
      & \textbf{$P(g_\text{true})$} & Brier Score & \textbf{$P(g_\text{true})$} & Brier Score & \textbf{$P(g_\text{true})$} & Brier Score \\
    \hline
    Humans (with instructions)  & 0.51 (0.06) & 0.10 (0.06) & 0.56 (0.21) & 0.10 (0.04) & 0.94 (0.08) &0.01 (0.01)\\
    Humans (without instructions)  & 0.23 (0.05) & 0.20 (0.03) & 0.44 (0.15) & 0.13 (0.05) & 0.92 (0.02) &0.00 (0.00)\\
    Model (with instructions) & 0.64 (0.23)  & 0.09 (0.06)  & 0.65 (0.23)  & 0.09 (0.06) & 0.99 (0.02) & 0.00 (0.00) \\
    Model (without instructions)  & 0.25 (0.00)  & 0.19 (0.00)  & 0.55 (0.17) & 0.11 (0.04) & 0.98 (0.04) & 0.00 (0.00) \\
    \hline
    \end{tabular}
    \vspace{0pt} 
    \caption{Goal inference metrics for humans (averaged across subjects) and our model across both experimental conditions. We report the probability assigned to the true goal $P(g_\text{true})$ and Brier score (lower is better) at the initial, median, and final judgment points $(t \in \{\text{first}, \text{median}, \text{last}\}$). Values are averaged over all stimuli, and standard deviations are in brackets.}
    \label{tab:goal_accuracy}
\end{table*}

\begin{figure}[h]
    \centering
    \includegraphics[width = 0.95\columnwidth]{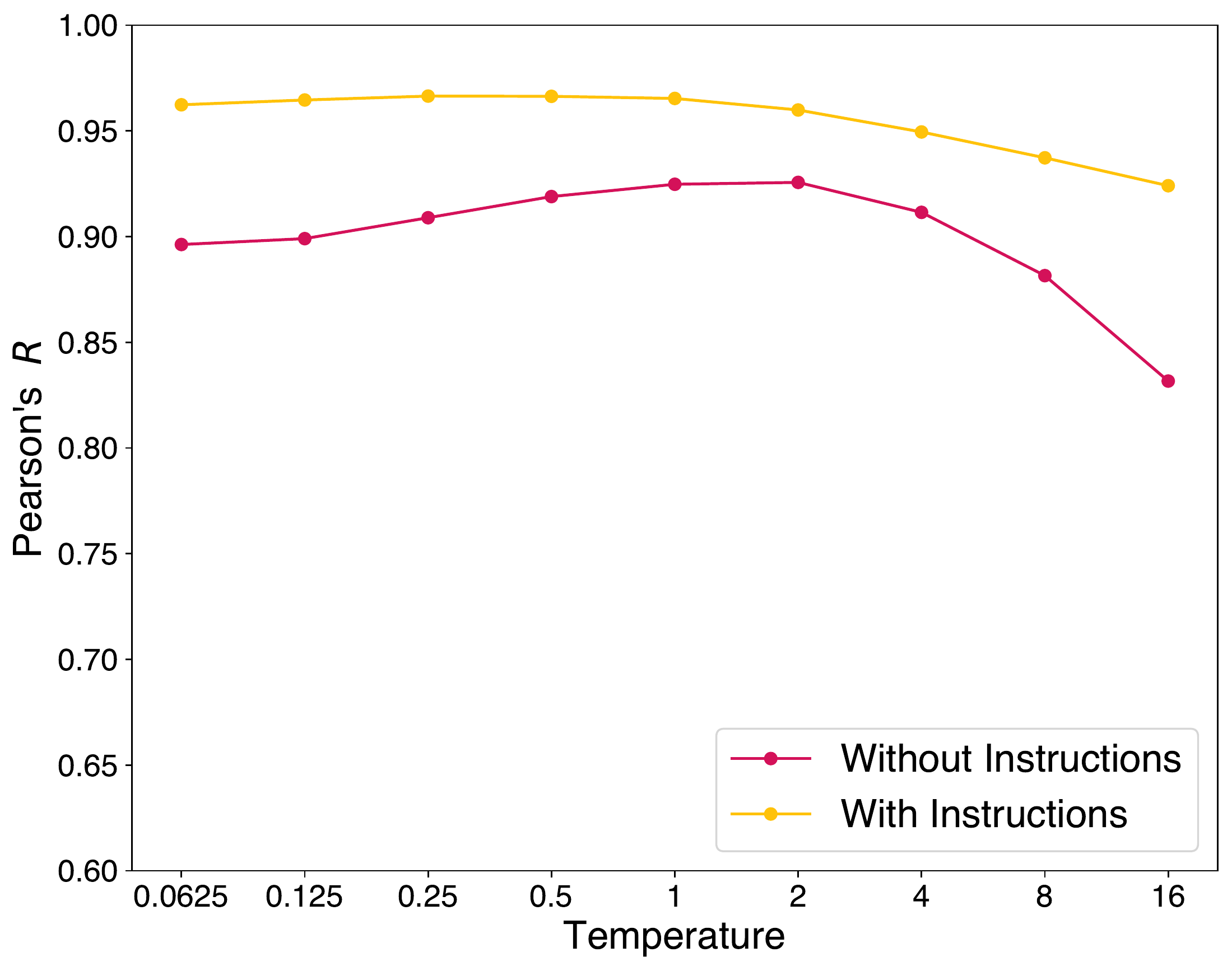}
    \caption{Correlation between model and human inferences under different model temperatures. In both conditions, the models show good fit but the correlation coefficients start to decline with temperature values greater than 2.}
    \label{fig:temperature}
    \vspace{-12pt}
\end{figure}

\paragraph{Sensitivity Analysis}
To evaluate the robustness of our correlational results to changes in model parameters, we also performed a sensitivity analysis, computing correlations for every value of $T$ we tested. Results of this analysis are shown in Figure \ref{fig:temperature}. We find that our model correlates highly with humans under many temperature settings, though correlations eventually decrease once temperature becomes sufficiently large. In the without-instructions condition, the correlation increases initially and peaks at temperature value of 2 before declining, while in the with-instructions condition, the performance starts to decline after a temperature value of 1. This suggests that humans tend to expect a small to moderate amount of action noise in this setting.

\paragraph{Goal Inference Accuracy}
Finally, we analyze the performance of both humans and our model in terms of accurately inferring the team's true goal. At each judgment point (which includes the initial state where no actions are shown), we compute the probability assigned to the true goal $P(g_\text{true})$ by both our model and the average human, along with the Brier score $\sum_{i} (P(g_i) - \mathbf{1}[g_i = g_\text{true}])^2$, a measure of well-calibrated inference. We then average these values across the entire dataset at each judgment point. We show these results for the initial, median, and final judgment points in Table \ref{tab:goal_accuracy}.  (In cases where a stimulus has no midpoint, we average the true goal probabilities for the middle two judgment points.)

We find that, on average, our model assigns slightly higher probability to the true goal than humans ($p<0.001$ in both experimental conditions via a paired $t$-test), indicating that our model is able to effectively infer the team's goals for the tested stimuli . Notably, there is a very clear difference between experimental conditions at the initial timestep: When instructions are provided at the start, both humans and the model assign much higher probability to the true goal than when no instructions are observed. This illustrates the importance of language in conveying useful information rapidly, as compared to inferring goals from actions alone.

Since instructions provide a lot of information, we also see that the true goal probability at the median judgment point only increases marginally in the with-instructions condition, with the actions observed between the first and median points providing limited extra information. In contrast, when only actions are observed, the increase in $P(g_\text{true})$ is much more pronounced. However, instructions alone are not enough to disambiguate the goal: As inference progresses on to the last judgment point, goal accuracy is significantly higher than at the first judgment point (corresponding to an "Instructions Only" baseline) and median judgment point.

Lastly, we analyze how instructions affect variance in goal inferences among our participants. By computing the standard deviation of at each judgment point among human participants, we observe a lower standard deviation in human ratings (SD = $0.150$) when instructions are provided, compared to the condition without instructions (SD = $0.188$), and find that this difference is statistically significant ($t=7.71$, $p<0.001$). This indicates that instructions do not only improve average accuracy, but also reduce individual variability when humans infer others' goals. 

\section{Related Work}

In addition to Bayesian theory-of-mind, rational speech act theory, and the Imagined We approach to cooperation, our model is closely related to several other lines of research:

\paragraph{Multimodal Goal Inference and Reward Learning.} Inferring goals can be framed as online inverse reinforcement learning (IRL), where the aim is to infer a reward function that explains an agent's behavior in a \emph{single} episode \cite{jara2019theory}. Our model is hence related to but distinct from IRL methods that learn reward functions from paired datasets of instructions and demonstrations \cite{williams2018learning,tung2018reward,fu2019language}. Most closely related is \emph{reward-rational implicit choice} \cite{jeon2020reward}, a framework for multimodal Bayesian reward learning from multiple types of human feedback, including demonstrations and language. Indeed our architecture can be viewed as a practical instantiation of this framework for those modalities, using our LLM utterance model (Figure \ref{fig:utterance-model}) as the (inverse) grounding function between utterances and trajectories.

\paragraph{Value Alignment and Assistance Games.} The principal-assistant team setting that we study is inspired by the formalization of human-AI value alignment as \emph{assistance games} \cite{hadfield2016cooperative}. While our focus here is on how an external observer would infer the goal of such a team, following the Imagined We approach, our group agent model could also be used by the assistant to infer the \emph{joint} goal and plan that the principal has in mind\footnote{With the modification that the assistant should not condition on their own actions to infer the shared goal.}. Our model can thus be seen as an alternate approach to practically solving assistance games, requiring less recursion than iterated best-response \cite{hadfield2016cooperative}, while avoiding the intractability of computing the pragmatic-pedagogical equilibrium \cite{fisac2017pragmatic, milli2017should}.

\paragraph{Instruction Following with Language Models.} Our work builds upon a long tradition of grounded instruction following from natural language \cite{tellex2011understanding}, leveraging an LLM utterance likelihood to achieve wide coverage over utterances without task-specific training. Whereas numerous papers have used LLMs to translate natural language to actions \cite{ahn2022can} or task specifications \cite{kwon2023reward,yu2023language,liu2022lang2ltl}, our approach is best viewed as a successor to Bayesian approaches such as \citet{squire2015grounding}, using LLMs in place of classical BoW models for natural language commands.

\section{Discussion and Future Work}

In this paper, we extended prior models of Bayesian goal inference to a multi-modal, multi-agent setting. Experiments demonstrate that our model can explain human inferences of team goals in a range of different cooperative scenarios, and that both our model and humans can effectively infer goals of cooperating agents in a multi-agent setting. Importantly, linguistic communication provides highly useful information that enables observers to more reliably infer a team's goal. As communication is ubiquitous in multi-agent contexts, our contributions provide a means to better modeling, understanding, and inferring shared plans and goals, which are crucial for applications in human-AI collaboration.

However, although our model shows human-like performance on our tested stimuli, it is still limited to relatively simple multi-agent scenarios. For instance, our model assume that agents follow a Boltzmann-rational policy, which does not account for bounded rationality \cite{alanqary2021modeling} or other systematic deviations from optimality \cite{shah2019feasibility}. As a result, we have anecdotally found that goal inference can be sensitive to some deviations from optimal behavior, such as the human simply staying idle for extra timesteps. In real-life scenarios, humans may not act optimally, or may have certain preferences when dividing tasks among team members (e.g. they may prefer minimizing their movements and assigning more work to the assistive agent). To address these issues, we aim to make our model more robust to boundedly-rational behavior, and to account for different team structures and preferences.

Along similar lines, our current model of communicative team utterances is at once somewhat heuristic and overly rational. The heuristic aspect is that we have manually defined which actions are salient in order to approximate pragmatic communication, but ideally this could be done in more principled manner that accounts for a wider diversity of utterances. For example, a more thoughtful speaker might come up with multiple plans that a listener might be expecting, and then communicate those actions that are \emph{most unique} to the plan that the speaker actually has in mind. The overly-rational aspect is that we have modeled utterances as communicating actions that come from \emph{only} the optimal lowest-cost plan. But real-world speakers are boundedly-rational at best, and may either accidentally omit salient actions from the plan, or communicate actions from a less optimal plan. Figuring out how to model these boundedly-rational speech acts is an important line of future work.

In sum, we are still some ways away from a complete model of rational communicative cooperation, not least due the many theoretical and technical challenges involved in defining and implementing reasonably optimal communicative behavior. Nonetheless, we have made a number of important steps in this direction by combining aspects of both Bayesian theory of mind and rational speech act theory into a combined model of sequential action and communication, while also using the power of probabilistic programming, model-based planning, and neural language models to go beyond the toy models that approaches like ours have previously been limited to. Using the technical infrastructure we have introduced in this paper, we look forward to addressing the challenges we have laid out above, and many more still.

\section*{Acknowledgements}

This work was funded in part by the DARPA Machine Common Sense, AFOSR, and ONR Science of AI programs, along with the MIT-IBM Watson AI Lab, and gifts from Reid Hoffman and the Siegel Family Foundation. Tan Zhi-Xuan is funded by an Open Phil AI Fellowship.

\bibliography{paper}
\bibliographystyle{tom2023}

\newpage
\appendix
\onecolumn

\section{Experiment Interface}

\begin{figure}[h]
    \centering
    \includegraphics[width=0.95\textwidth]{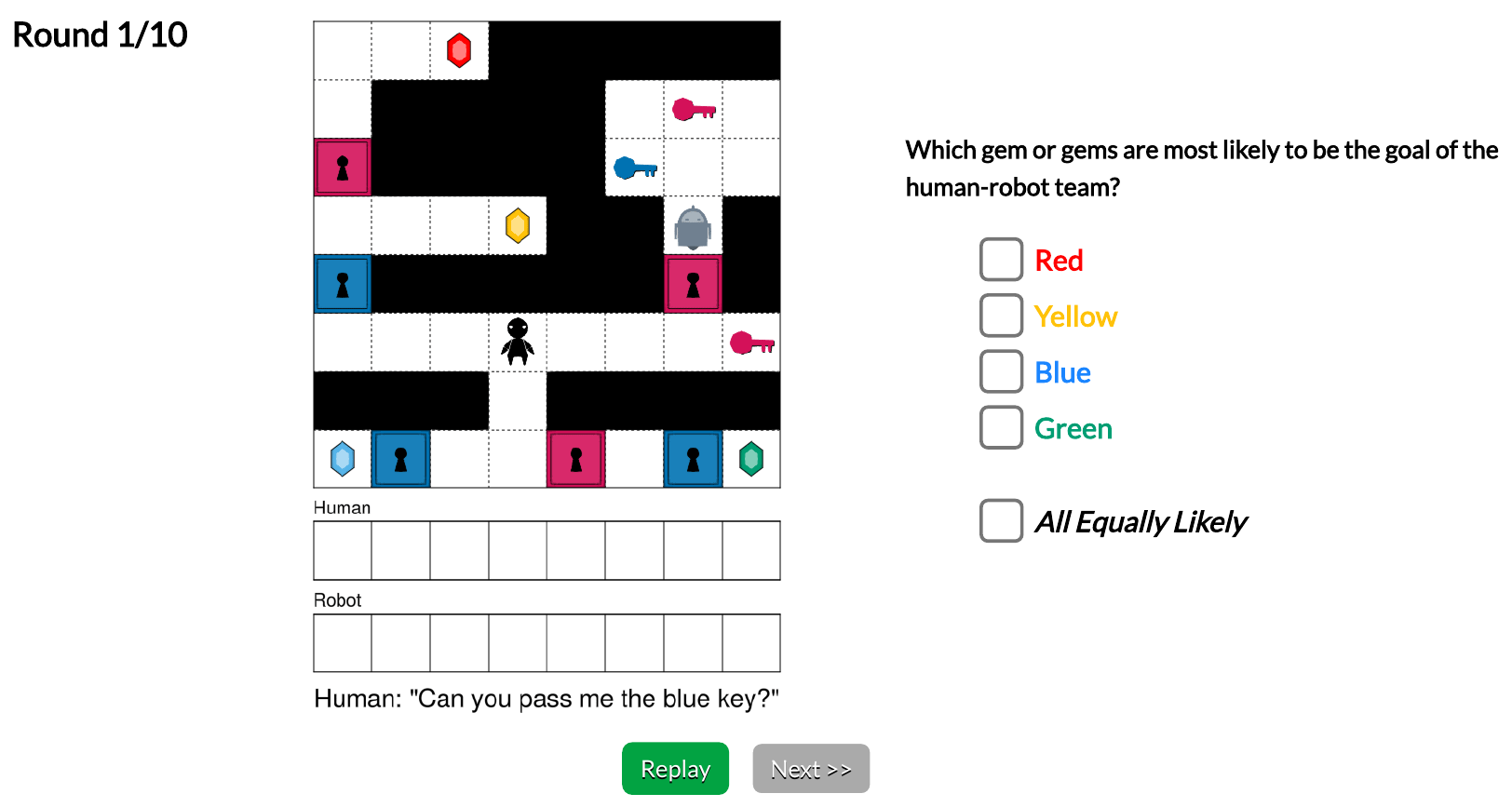}
    \caption{Interface for the "with instructions" condition of our human experiments.}
    \label{fig:interface}
\end{figure}

Figure \ref{fig:interface} shows the interface for our human experiments, which is adapted from \citet{alanqary2021modeling}. The interface displays animated stimuli which pause at specific judgment points. At each judgment point, participants provide goal inferences by selecting the gems(s) they believe to be the team's most likely goal, then proceed to next segment. Responses are converted to uniform distributions over the selected goals (e.g., if three goals were selected, 33.3\% probability is allocated to each goal). If no goal seems more likely than the others, participants can choose the \textit{All Equally Likely} option.  

\section{Experiment Stimuli and Additional Results}

In the supplementary information, we provide both storyboard plots and animated GIFs showing the 20 stimuli we used in our human experiment. We also provide goal inference storyboards over time for each stimulus, similar to Figure \ref{fig:storyboard}. This information can also be accessed at \url{https://osf.io/gh758/}.


\end{document}